\documentclass[a4paper]{article}

\usepackage{INTERSPEECH2019}
\usepackage{array}
\usepackage{graphicx} 
\usepackage{multirow}
\usepackage{color, soul}
\usepackage{url}
\usepackage{xcolor}
\usepackage{cite}
\usepackage[colorlinks,linkcolor=blue]{hyperref}
\usepackage{multirow}
\usepackage{amsmath}
\usepackage{bm}
\usepackage{graphicx}
\usepackage{comment}
\usepackage{longtable}
\usepackage{xparse}
\usepackage{booktabs}
\usepackage{subfigure}
 
\usepackage{array}
\usepackage{ulem}
\usepackage[ruled,vlined]{algorithm2e}
\usepackage{setspace}

\title{Reinforcement Learning for Emotional Text-to-Speech Synthesis \\ with Improved Emotion Discriminability }

\name{Rui Liu$^{1,2}$, Berrak Sisman$^{1}$, Haizhou Li$^{2,3}$ \thanks{Speech samples: \textcolor{magenta}{\href{https://ttslr.github.io/i-ETTS}{https://ttslr.github.io/i-ETTS}}. }}
 
\address{
$^1$ Singapore University of Technology and Design (SUTD), Singapore \\
 $^2$ ECE Department, National University of Singapore (NUS), Singapore\\
 $^3$ Machine Listening Lab, University of Bremen, Germany}
\email{liurui\_imu@163.com, berrak\_sisman@sutd.edu.sg, haizhou.li@nus.edu.sg}

\begin{document}

\maketitle

\begin{abstract}
Emotional text-to-speech synthesis (ETTS) has seen much progress in recent years. However, the generated voice is often not perceptually identifiable by its intended emotion category. To address this problem, we propose a new interactive training paradigm for ETTS, denoted as \textit{i-ETTS}, which seeks to directly improve the emotion discriminability by interacting with a speech emotion recognition (SER) model.
Moreover, we formulate an iterative training strategy with reinforcement learning to ensure the quality of \textit{i-ETTS} optimization.  
Experimental results demonstrate that the proposed \textit{i-ETTS} outperforms the state-of-the-art baselines by rendering speech with more accurate emotion style. To our best knowledge, this is the first study of reinforcement learning in emotional text-to-speech synthesis. 
\end{abstract}
\noindent\textbf{Index Terms}: Reinforcement Learning, Emotional Text-to-Speech Synthesis, Speech Emotion Recognition.

\section{Introduction}
Emotional text-to-speech (ETTS) seeks to synthesize human-like natural-sounding voice for a given input text with desired emotional expression. The recent advances have enabled many applications such as virtual assistants, call centers, dubbing of movies and games, audiobook narration, and online education. 
 
The early studies of ETTS are based on hidden Markov models~\cite{tokuda2002hmm, yamagishi2003modeling, eyben2012unsupervised}. For example, we can synthesize speech with a desired emotion through model interpolation~\cite{yamagishi2003modeling} or by incorporating unsupervised expression cluster during training ~\cite{eyben2012unsupervised}. 
Recently, deep learning opens up many possibilities for ETTS \cite{lorenzo2018investigating,choi2019multi}, where emotion codes can be used as control vectors to change text-to-speech (TTS) output in different ways. Successful attempts include global style tokens (GST) that have been used to control the expressiveness in emotional TTS ~\cite{hsu2018hierarchical,zhang2019learning,um2020emotional}. Such approaches typically use style embedding to indicate the emotion rendering and can learn speech variations 
\textcolor{black}{in an unsupervised manner}. We note that the latent style embeddings \textcolor{black}{has no explicit meaning}, hence lacks interpretability. \textcolor{black}{Therefore,} GST-Tacotron studies often face emotion confusion problem \cite{el2011survey}, i.e., the projected emotion in the synthetic speech is not rendered accurately.

Recently, some approaches \cite{wu2019end,cai2020emotion,li2021controllable} were proposed to enhance the interpretability of the style embedding. These approaches add an additional emotion recognition loss \cite{wu2019end,cai2020emotion} or perceptual loss \cite{li2021controllable} to force the latent style embedding to pay more attention to the emotion rendering. These works have made a great contribution to the development of ETTS. However, they mostly focus on the hidden features \cite{liu2020expressive}, not the explicit output; and optimize the output acoustic features with mean square error (MSE) loss, which calculates the distance between the synthesized emotional speech and natural reference. We note that such MSE-based objective function does not focus directly on emotion discriminability, hence often face emotion confusion problem in generated speech.

We are motivated by the fact that the interaction of speech emotion recognition model with ETTS can improve the emotion discriminability of synthesized speech, hence overcome emotion confusion problem. We note that ETTS can be trained with SER classification result in a supervised manner. However, this may require a large amount of emotion-labeled speech data, that limits the scope of applications. Therefore, this paper studies the use of reinforcement learning for ETTS in an interactive manner for improved emotion discriminability.

The reinforcement learning (RL) algorithm \cite{sutton2018reinforcement} learns how to achieve a complex goal in an interactive manner. Specifically, RL involves agents to learn their behavior by trial and error \cite{sutton2018reinforcement}. RL agents aim to learn decision-making by successfully interacting with the environment where they operate.
It has enabled speech processing systems \cite{latif2021survey} through a well designed feedback that reflects appropriate perceptual metrics in speech enhancement \cite{8369109,8683648}, speech recognition \cite{8462656} and speaker recognition \cite{seurin2020machine}. We note that the use of interactive paradigm with RL algorithm in emotional TTS remains to be explored, which will be the focus of this paper.
 
In this paper, we propose a novel ETTS framework with reinforcement learning, denoted as \textit{i-ETTS}. The proposed approach aims to overcome emotion confusion problem of synthesized speech by optimizing the ETTS model through an interaction with a SER model. The proposed idea is expected to reduce the requirement of emotion-labeled training data size.
Experimental results demonstrate that i-ETTS consistently outperforms the state-of-the-art baselines, by achieving improved emotion discriminability of synthesized speech. 
 
\begin{figure*}[t]
    \centering
    \vspace{-4mm}
    \centerline{\includegraphics[width=0.75\linewidth]{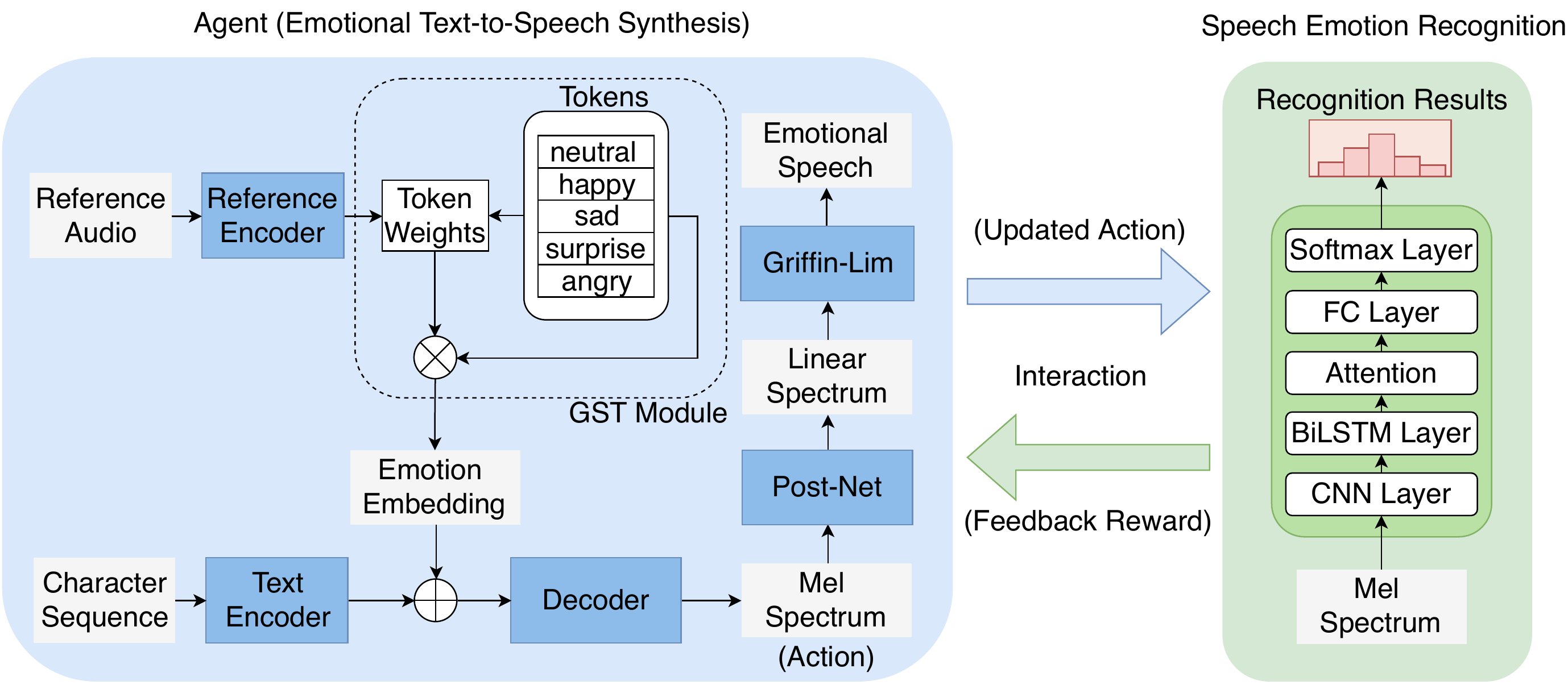}}
    \vspace{-2mm}
    \caption{Block diagram of the proposed interactive paradigm for emotional text-to-speech synthesis with reinforcement learning.}
    \label{fig:overall}
    \vspace{-5mm}
\end{figure*}

The main contributions of this paper are summarized as follows: 
1) we introduce a novel interactive emotional text-to-speech synthesis paradigm that overcomes emotion confusion problem;
2) we formulate the training problem with RL, and optimize the ETTS model with policy gradient and a reward function correlated with the SER accuracy;
3) we utilize an iterative training strategy for a stable training procedure. To our best knowledge, this is the first study of reinforcement learning in emotional speech synthesis.

This paper is organized as follows. We motivate our study through the comparison with existing RL and ETTS frameworks in Section \ref{sec:rela}. We formulate the proposed paradigm in Section \ref{model}. We report the results of a systematic evaluation and comparison in Sections \ref{exp}. Finally, Section \ref{con} concludes the study.

\vspace{-2mm}
\section{Related Work}
\label{sec:rela}
\vspace{-2mm}

The reinforcement learning (RL) algorithm takes an action in an environment in order to maximize the cumulative reward \cite{sutton2018reinforcement}. It has been successfully applied to various tasks, such as computer game~\cite{mnih2015human}, natural language processing~\cite{Luo19DualRL}, and speech processing~\cite{latif2021survey,8683648,8369109,8462656,seurin2020machine}.  For example, RL has been used to optimize speech enhancement system based on character recognition error. Moreover, it has also been used for DNN-based source enhancement \cite{8369109} where the objective sound quality assessment scores are given as the reward. Last but not least, RL-based methods have also been proposed for speech recognition \cite{8462656} and speaker recognition \cite{seurin2020machine}. We note that these approaches utilize limited or unlabeled data and optimize the target model by interacting with the feedback, which correlates with an evaluation metric from a third party directly.

With the advent of deep learning, end-to-end TTS systems, such as Tacotron \cite{wang2017tacotron}, Tacotron2 \cite{shen2018natural} and their variants \cite{wang2018style,ICASSP2020liu} greatly improve the voice quality of the synthesized speech. However, it remains a challenge as to how we generate speech with an intended emotion due to the fact that it is hard to interpret the learned style embedding.

Some recent studies make use of SER loss \cite{wu2019end,cai2020emotion} and perceptual loss \cite{li2021controllable} to assign specific emotional information to the style embedding. 
However, the commonly used MSE loss for output feature does not directly improve the perceptual discriminability of the synthesized speech.
In this paper, we devise a training strategy under reinforcement learning, which employs an interactive game between the ETTS system and a SER to directly improve emotion discriminability.

\vspace{-2mm}
\section{Interactive ETTS: Methodology}
\label{model}
\vspace{-2mm}
We propose a novel interactive training paradigm for emotional TTS  under the reinforcement learning framework, denoted as \textit{i-ETTS}.
Our method is a combination of ETTS and RL, hence we first present the main components of proposed i-ETTS that includes \textit{agent}, \textit{policy}, \textit{action} and \textit{reward}. We then introduce the action update method, denoted as policy gradient. Finally, we describe the iterative training algorithm.

\vspace{-3mm}
\subsection{i-ETTS: Interactive ETTS with RL}
\vspace{-2mm}
As in Fig. \ref{fig:overall}, the ETTS model can be viewed as an \textit{agent} under the reinforcement learning framework. The parameters of this agent define a \textit{policy}.
\textcolor{black}{The policy aims to predict the emotional acoustic features at each time step. The emotional acoustic features define the \textit{action}. After the emotional acoustic feature prediction is finished, the pre-trained SER starts to feedback the emotion recognition accuracy, which is denoted as \textit{reward}. The policy gradient strategy was used to perform back-propagation smoothly, and optimize the ETTS model to achieves maximum reward.}

\vspace{-3mm}
\subsubsection{Agent: ETTS}
\vspace{-2mm}
As shown in the left panel of Fig. \ref{fig:overall}, our emotional text-to-speech synthesis model is built on the GST-based Tacotron network \cite{wang2018style}. It consists of text encoder, reference encoder with a GST module attached, attention-based decoder and Post-Net. We use Griffin-Lim algorithm \cite{Griffin1984Signal} to reconstruct the speech waveform.

Text encoder shares a similar architecture with that of Tacotron2~\cite{shen2018natural}, which takes the character embeddings as input and generates high-level encoder output embeddings.
The reference encoder and GST module were defined to encode the emotion of reference audio into a fixed-length emotion embedding. Their architectures are also similar to that of \cite{wang2018style}, except for the definition of style tokens. To force the GST module to pay more attention to learn the emotion-related style, \textcolor{black}{we set the number of tokens to the number of emotion categories in the corpus.}
The attention-based decoder takes the encoder output and emotion embedding as input and predicts the acoustic features frame-by-frame.
Note that we also adopt a similar decoder structure to Tacotron2 \cite{wang2018style}, except that the post-net as in \cite{wang2017tacotron} was added to convert the mel spectrum to linear spectrum. The GMM attention \cite{graves2013generating} was utilized to learn the $<$text, wav$>$ alignment.

\vspace{-3mm}
\subsubsection{Reward: SER Accuracy}
\label{sec:ser}
\vspace{-2mm}
In order to achieve emotion expressiveness and discriminability, we introduce a reward function correlated with the SER evaluation metric. 

As shown in the right panel of Fig. \ref{fig:overall}, 
the SER network includes a CNN layer, a BiLSTM layer, an attention layer, a fully connected (FC) layer and a softmax layer. The above architecture is proven effective to learn discriminative features for SER \cite{chen20183}.
The CNN first reads a mel spectrum from input utterance and then outputs a fixed size latent representation. The BiLSTM summarizes the temporal information into another latent representation. The attention layer learns the weights for each frame. The final softmax layer outputs the probability of five emotion types, i.e., happy, angry, neutral, sad and surprise.

For ETTS training, given an input text $\boldsymbol{x_{i}}$ and reference audio $\boldsymbol{\hat{y}_{i}}$ with emotion label $\boldsymbol{l_{\hat{y}_{i}}}$, the goal is to generate an emotional speech $\boldsymbol{y^{\prime}_{i}}$ that is not only natural, but also in accordance with the desired emotion $\boldsymbol{l_{\hat{y}_{i}}}$.
{
One way to employ the pre-trained SER as an expert system in place of a human judge is to evaluate the emotion category of the generated speech $\boldsymbol{y^{\prime}_{i}}$. We then use the recognition accuracy as the reward to update the ETTS model.}

Specifically, upon generating the end-of-sequence (EOS) token, the pre-trained SER is used to evaluate how well the generated mel spectrum feature $\boldsymbol{y^{\prime}_{i}}$ matches the reference emotion label $\boldsymbol{l_{\hat{y}_{i}}}$.
Mathematically, the recognition probability $\boldsymbol{p_{i}}$ about the target emotion $\boldsymbol{l_{\hat{y}_{i}}}$ of $\boldsymbol{y^{\prime}_{i}}$ is formulated as:
\vspace{-2mm}
\begin{equation}
\boldsymbol{p_{i}} = {\rm SER} \left(\boldsymbol{l_{\hat{y}_{i}}} \mid \boldsymbol{y}^{\prime}_{i} ; \boldsymbol{\varphi}\right) 
\vspace{-2mm}
\label{eq:SER}
\end{equation}

\noindent{where $\boldsymbol{\varphi}$ represents all the parameters of SER, that are pre-trained before the training of ETTS. Note that the value of $\boldsymbol{p_{i}}$ ranges from 0 to 1.}

{
During ETTS training, we sample a fixed number of predicted mel spectrum features $\boldsymbol{y}^{\prime}$ from a batch of data to calculate the SER reward, which is formulated as:
\vspace{-2mm}
\begin{equation}
R =  \frac{N}{K} =  \frac{\sum_{i=1}^{K} \boldsymbol{1} {(\boldsymbol{p_{i}} > \lambda)}}{K}
\vspace{-2mm}
\label{eq:reward}
\end{equation}
where $R$ is the reward of the sampled mel spectrum features $\boldsymbol{y}^{\prime}$ from ETTS model. Note that the value of reward $R$ also ranges from 0 to 1. $K$ denotes the sample size.
$\lambda$ is a threshold that is set to 0.5 in this work. $N$ represents the number of samples of which the probability $\boldsymbol{p_{i}}$ of the target emotion exceeds the threshold $\lambda$. $\boldsymbol{1}$($\cdot$) is an indicator function of a set $\{0,1\}$.
}


\subsubsection{Action update: Policy Gradient}

We use policy gradient algorithm \cite{williams1992simple} to estimate the gradient that leads to a larger expected reward $\mathbb{E}[R]$ of the generated acoustic features $\boldsymbol{y}^{\prime}$ for input text $x$. 
Note that the gradient of the ETTS model $P(\cdot)$ w.r.t. the model parameters $\boldsymbol{\theta}$ is estimated by sampling as follows:
 
\vspace{-3mm}
\begin{equation}
\begin{aligned}
\nabla_{\boldsymbol{\theta}} \mathbb{E}[R] &=\nabla_{\boldsymbol{\theta}}  P\left(\boldsymbol{y}^{\prime} \mid \boldsymbol{x} ; \boldsymbol{\theta}\right) R \\
&= P\left(\boldsymbol{y}^{\prime} \mid \boldsymbol{x} ; \boldsymbol{\theta}\right) {R} \!\!\!\!\!\quad \nabla_{\boldsymbol{\theta}} \log \left(P\left(\boldsymbol{y}^{\prime} \mid \boldsymbol{x} ; \boldsymbol{\theta}\right)\right) \\
& \simeq   R  \!\!\!\!\!\quad \nabla_{\boldsymbol{\theta}} \log \left(P\left(\boldsymbol{y}^{\prime} \mid \boldsymbol{x} ; \boldsymbol{\theta}\right)\right)
\vspace{-2mm}
\end{aligned} 
\label{eq:update}
\end{equation}

Unlike the traditional gradient descent algorithm, the policy gradient algorithm estimates the weights of an optimal policy through gradient ascent. Note that it can assign an explicit emotion-aware supervised signal to the gradient during the model training, resulting in emotional speech with accurate emotion category.

\subsection{Iterative Training}

In practice, it is hard to train the whole network from scratch with policy gradient since the ETTS model may find an unexpected way to achieve a high reward but fail to guarantee the naturalness or robustness of the synthesized speech \cite{Luo19DualRL}. Therefore, we formulate an iterative training algorithm, reported in Algorithm~\ref{algo:main}, that contains two main steps: 1) pre-training, and 2) iterative-training. 
First, we use \textit{MSE Loss} to pre-train the ETTS model with text-wav pairs from the training set. After the pre-training phase, iterative-training phase aims to optimize the ETTS model with reward and the MSE loss alternatively.

\begin{algorithm}[t]
\setstretch{0.9}
\SetAlgoLined
\textbf{Input}: Training set: $D = \{x ,y ,l \} $\\
\quad \quad $x $: character sequence\\
\quad \quad $y $: acoustic feature sequence\\
\quad \quad $l $: emotion label\\
\textbf{Output}: ETTS model: $\boldsymbol{\theta}$ \\
\textbf{Begin}
\\0. Pre-train SER model $\boldsymbol{\varphi}$ with $(y, l)$
\\ $\triangleright$ \textbf{Pre-training}
\\1: Pre-train ETTS model $\boldsymbol{\theta}$ using \textit{MSE Loss} with $(x,y,l)$
\\ $\triangleright$ \textbf{Iterative-training}
\\3: \textbf{for} epoch = 1,2,...N \textbf{do}
\\4. \quad Sample a batch $B$ in $D$
\\5. \quad Sample $(x, y, l)$ from $B$
\\6. \quad Generate mel spectrum features $y^{\prime}$ via $\theta$
\\7. \quad Compute reward $R$ based on Eq. \ref{eq:reward}
\\8. \quad Update $\theta$ using $R$ based on Eq. \ref{eq:update}
\\9. \quad Update $\theta$ using \textit{MSE Loss} with $(x,y,l)$
\\10: \textbf{end for}
\\\textbf{End}
\caption{\!Iterative Training Algorithm.\!\!}
\label{algo:main}
\end{algorithm}

\section{Experiments}
\label{exp}
We report emotional TTS experiments on ESD database \cite{zhou2020seen}, which is a new publicly available emotional speech dataset for emotional speech synthesis. ESD is a multi-lingual dataset and has 350 parallel utterances spoken by 10 native English and 10 native Mandarin speakers. We use the English corpora with a total of nearly 13 hours of speech by 5 male and 5 female speakers in five emotions, namely happy, angry, neutral, sad and surprise. In ESD, scripts are provided, speech data are sampled at 16 kHz and coded in 16 bits. We note that ESD is considered to be large enough for various voice conversion tasks \cite{sisman2020overview}, while it provides limited data for TTS and ETTS tasks.

\subsection{Comparative Study}
We implement two baseline systems together with the proposed \textit{i-ETTS}, as summarized next.
\begin{itemize}
    \item MTL-ETTS \cite{cai2020emotion}:~An emotional TTS model that jointly trains an auxiliary SER task with the TTS model;
    \item CET-ETTS \cite{li2021controllable}:~An emotional TTS model that uses two reference encoders with SER module and perceptual loss to enhance the emotion-discriminative ability;
    \item i-ETTS: the proposed \textit{i-ETTS} that optimizes the ETTS model with a reward function correlated with the SER accuracy. 
\end{itemize}

For a fair comparison, all frameworks use the same SER module, as also illustrated in Section \ref{sec:ser}, and  Griffin-Lim algorithm \cite{Griffin1984Signal} is used for waveform generation. We note that the use of neural vocoder will further improve the speech quality \cite{zhou2020multi}, which is not the main focus of this paper.

\begin{figure*} [th!]
\centering
\begin{minipage}{\linewidth}
  \centerline{
  \includegraphics[width=0.21\linewidth]{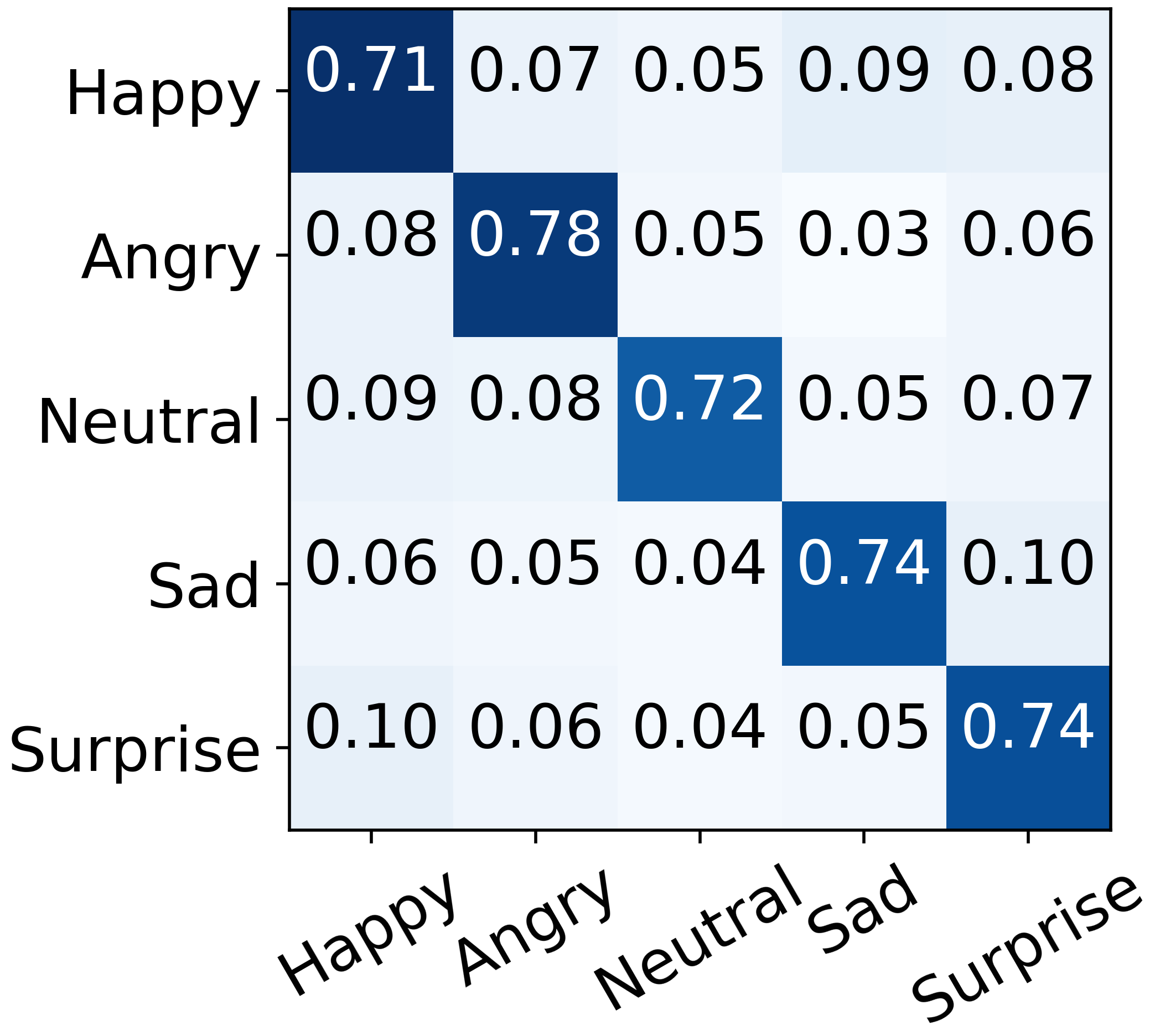}
  \includegraphics[width=0.209\linewidth]{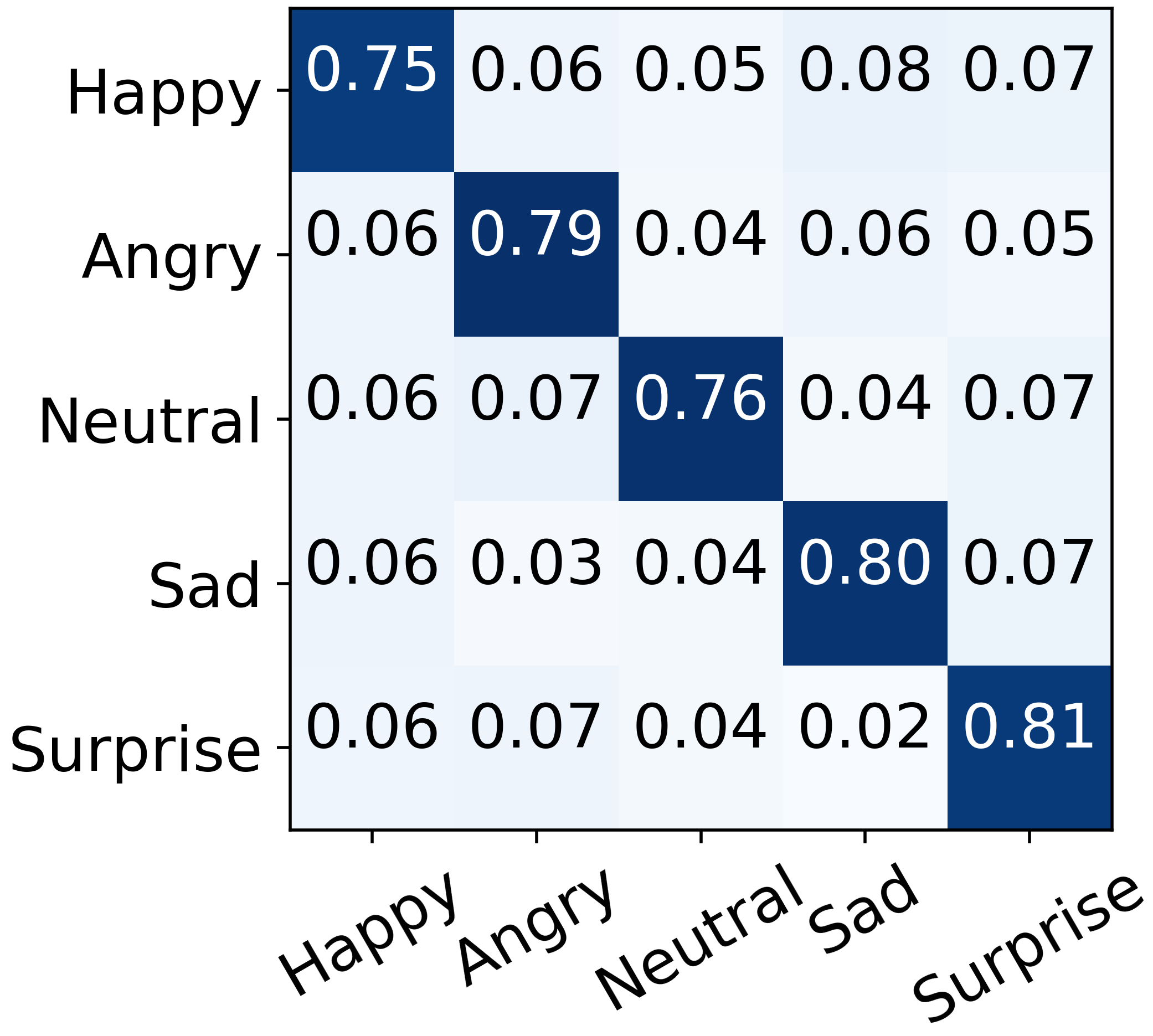}
  \includegraphics[width=0.245\linewidth]{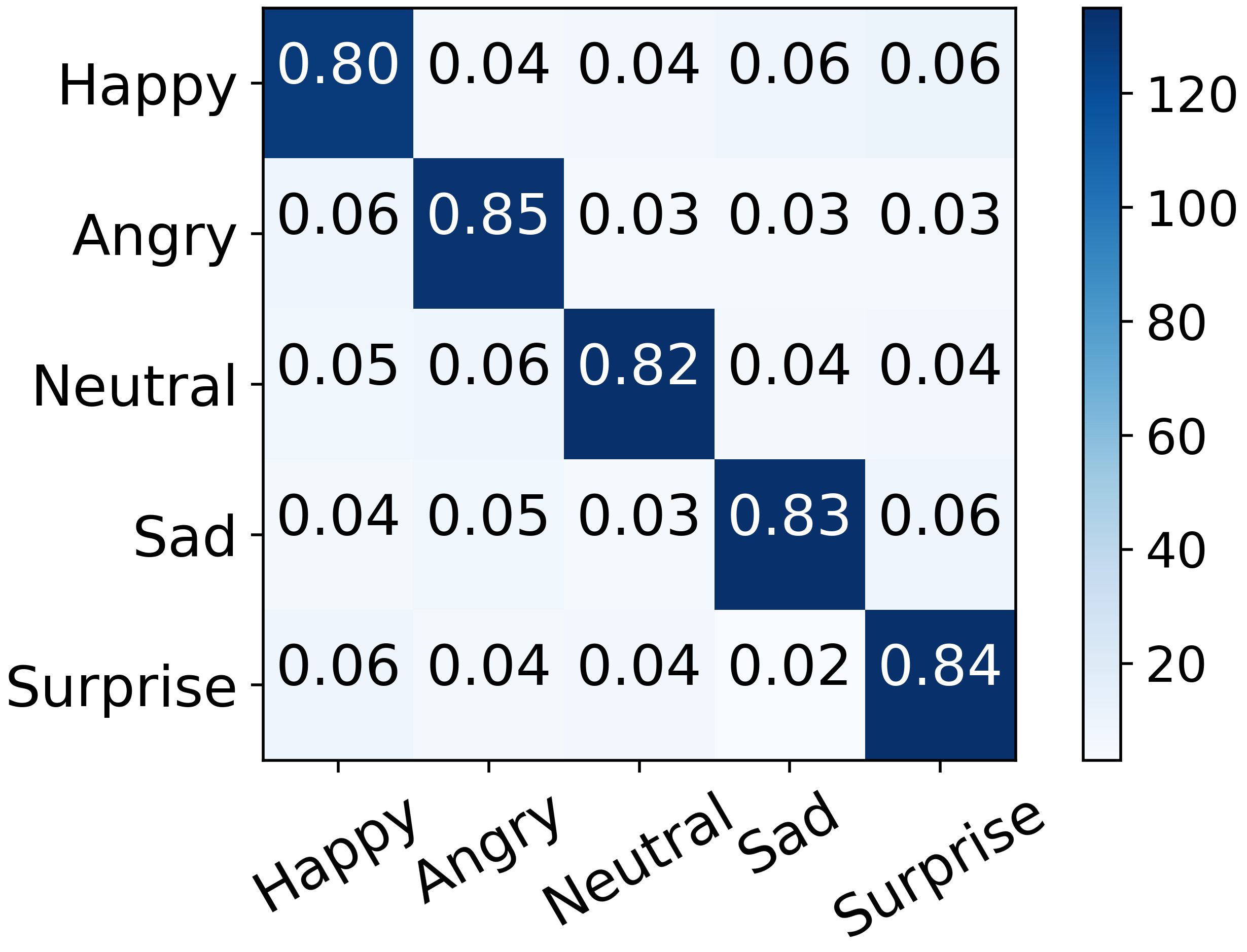}
  }
  \centerline{
  (a) MTL-ETTS \quad\quad\quad\quad\quad\quad 
  (b) CET-ETTS \quad\quad\quad\quad\quad\quad 
  (c) i-ETTS
  }
\end{minipage}
\vspace{-3mm}
\caption{Confusion matrices of synthesized speech from MTL-ETTS, CET-ETTS and the proposed i-ETTS. The X-axis and Y-axis of subfigures represent predicted and truth emotion label, respectively.}
\vspace{-3mm}
\label{fig:SER}
\end{figure*}

\begin{figure}[t]
    \centering
    \centerline{\includegraphics[width=0.88\linewidth]{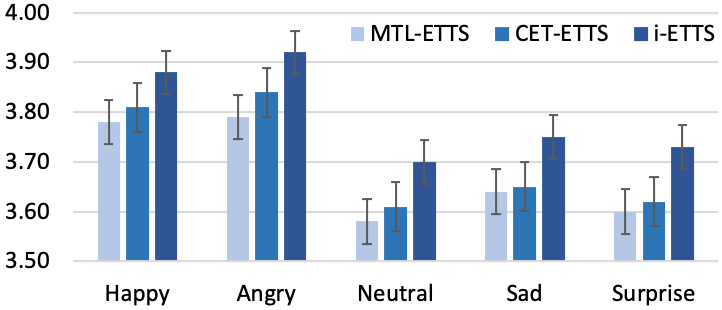}}
    \vspace{-2mm}
    \caption{The MOS test results of MTL-ETTS, CET-ETTS and i-ETTS for five emotion categories, with 95\% confidence interval.}
    \label{fig:MOS}
    \vspace{-5mm}
\end{figure}

\subsection{Experimental Setup}

For all systems, the encoder takes the character sequence as input. The 80-channel mel spectrum and 1025-channel linear-spectrum are extracted with 12.5ms frame shift and 50ms frame length.
The decoder predicts 5 output frames at each decoding step.
We randomly split the audio set of each emotion in each speaker into training/validation/test set with the number of 300/30/20, resulting in a total of 15,000/1,500/1,000 training/validation/test proportion for whole dataset. 
\textcolor{black}{We use the Adam optimizer with $\beta_{1}$ = 0.9, $\beta_{2}$ = 0.999 to optimize the model. 
The learning rate is set to $10^{-3}$ before 100k steps, then exponentially decays to $10^{-5}$ after 100k steps. 
We set the batch size to 32. The sample size in Eq. \ref{eq:update} is set to 20.}
The MTL-ETTS and CET-ETTS systems are trained with 300k steps.

For our i-ETTS, we first pre-train a SER \cite{chen20183}, that reports a   classification accuracy of 90.4\% for all emotions on the test set. During the iterative learning, i-ETTS model is first pre-trained 200k steps via MLE, then the iterative training runs until the performance on validation set converges. Note that we set the number of tokens in GST module to 5 to be consistent with the number of emotion categories in ESD dataset. The GST module produces a 256-dimensional emotion embedding. At run-time, the task is to generate emotional speech by selecting a reference audio with the desired style. In this paper, we follow \cite{kwon2019effective} and manually specify the weight of style tokens by averaging the style token weights of the test set for each emotion. All systems use Griffin-Lim algorithm \cite{Griffin1984Signal} with 64 iterations.

\subsection{Experiment Results}
\subsubsection{Objective Evaluation}
We first evaluate the emotion-discriminative ability of the synthesized speech with SER accuracy among the baselines MTL-ETTS, CET-ETTS and the proposed i-ETTS. As reported in Fig. \ref{fig:SER}, the proposed \textit{i-ETTS} consistently outperforms two baselines by achieving average accuracy of 82.8\%, which is much higher than that of the MTL-ETTS (73.8\%) and CET-ETTS (78.2\%). The results show the effectiveness of our proposed model in terms of emotion expressiveness and discriminability.

\subsubsection{Subjective Evaluation}
 \vspace{-1mm}
\textcolor{black}{We first conduct mean opinion score (MOS) listening experiments to compare overall emotion expression among different frameworks.}
The MOS values are calculated by taking the arithmetic average of all scores provided by the subjects. We keep the linguistic content same among different models to exclude other interference factors.  \textcolor{black}{We invite 15 subjects to participate in these experiments. Each listener listened to 100 synthesized speech samples.} As shown in Fig. \ref{fig:MOS}, we observe that the proposed \textit{i-ETTS} achieves remarkable results by outperforming both MTL-ETTS and CET-ETTS baseline systems for all emotion categories.

\begin{figure}[t]
    \centering
    \vspace{1mm}
    \centerline{\includegraphics[width=\linewidth]{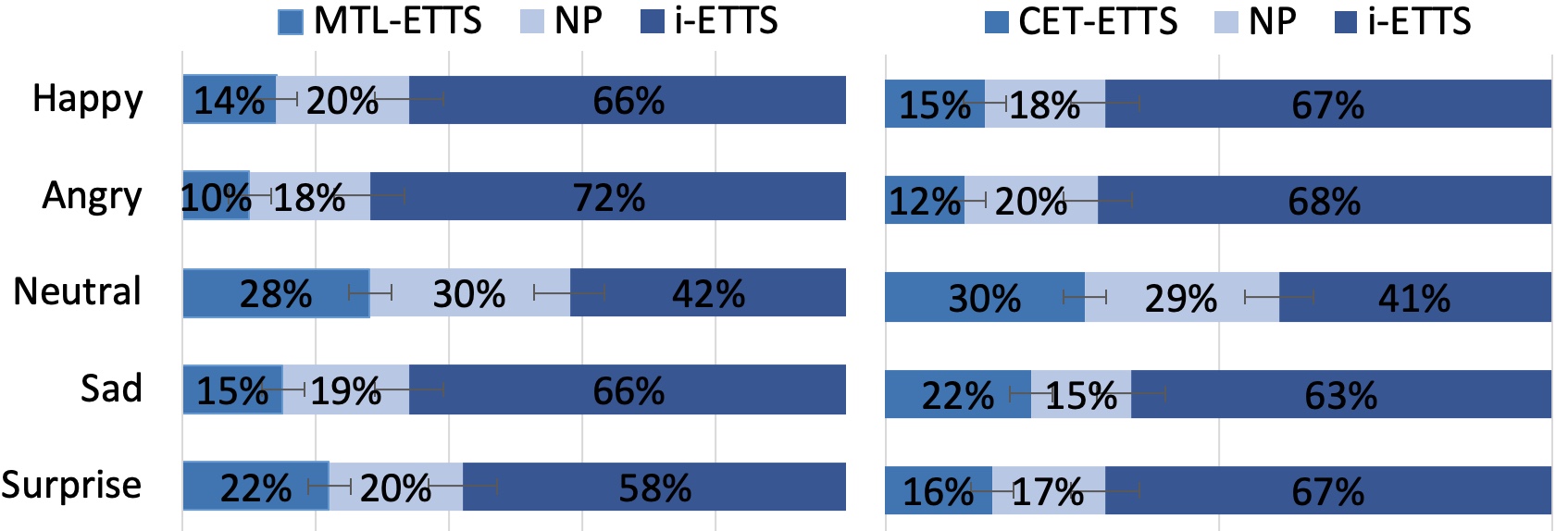}}
     \vspace{-3mm}
    \caption{The preference test results between 1) MTL-ETTS $vs.$ i-ETTS, and 2) CET-ETTS $vs.$ i-ETTS for five emotion categories, with 95\% confidence interval.}
    \vspace{-5mm}
    \label{fig:AB}
\end{figure}

\textcolor{black}{We also conduct A/B preference experiments as reported in \ref{fig:AB} to evaluate the emotion expressiveness} among MTL-ETTS, CET-ETTS and \textit{i-ETTS} systems. \textcolor{black}{We ask listeners to listen to all speech samples from different systems, and then choose the preferred one in terms of emotion expression. We also invite 15 subjects to this experiment. Each listener listened to 100 synthesized speech samples.}
Consistent with the previous experiments, the preference test results in Fig. \ref{fig:AB} show that our proposed \textit{i-ETTS} system can generate more expressive emotional speech, and always achieves better results than MTL-ETTS and CET-ETTS systems in all five emotion categories. All the above observations validate the effectiveness of our proposed \textit{i-ETTS} system in terms of emotion expressiveness and discriminability.

\vspace{-4mm}
\section{Conclusion}
\label{con}
We have studied an interactive training paradigm for emotional TTS, denoted as \textit{i-ETTS}, to synthesize emotional speech with accurate emotion category.
In doing so, we devise a training strategy under reinforcement learning, which employs an interactive game between ETTS and SER. We formulate a policy gradient strategy and a reward function correlated with the SER accuracy. A series of experiments were conducted to evaluate the emotion expression. The proposed \textit{i-ETTS} achieves remarkable performance by consistently outperforming the ETTS baseline systems in terms of voice quality and emotion discriminability. Future work includes improving the performance by investigating more effective ways to optimize the ETTS model.

\vspace{-2mm}
\section{Acknowledgements}
The research is funded by SUTD Start-up Grant Artificial Intelligence for Human Voice Conversion (SRG ISTD 2020 158) and SUTD AI Project (SGPAIRS1821) Discovery by AI - The Understanding and Synthesis of Expressive Speech by AI. This research by Haizhou Li supported by the Science and Engineering Research Council, Agency of Science, Technology and Research, Singapore, through the National Robotics Program under Grant No. 192 25 00054 and Programmatic Grant No. A18A2b0046 from the Singapore Government’s Research, Innovation and Enterprise 2020 plan (Advanced Manufacturing and Engineering domain). Project Title: Human Robot Collaborative AI for AME.


\normalem 
\bibliographystyle{IEEEtran}
\bibliography{refs}

\end{document}